\documentclass[11pt,a4paper]{article}
\usepackage[T1]{fontenc}
\usepackage[utf8]{inputenc}
\usepackage{lmodern}
\input{glyphtounicode}
\pdfglyphtounicode{parenleftbig}{0028}
\pdfglyphtounicode{parenrightbig}{0029}
\pdfglyphtounicode{bracketleftbig}{005B}
\pdfglyphtounicode{bracketrightbig}{005D}
\pdfglyphtounicode{summationtext}{2211}
\pdfglyphtounicode{summationdisplay}{2211}
\usepackage[margin=1.1in,nohead,includefoot,footskip=24pt]{geometry}
\usepackage{amsmath,amssymb}
\usepackage{graphicx}
\usepackage{booktabs,multirow}
\usepackage{algorithm}
\usepackage{algorithmic}
\usepackage[protrusion=false]{microtype}
\usepackage[numbers,sort&compress]{natbib}
\usepackage{caption}
\usepackage{titlesec}
\usepackage{hyperref}
\hypersetup{
  hidelinks,
  pdftitle={DDSL: Iterative Self-Labeling for Financial Market States},
  pdfauthor={Xianbin Gu; Qingyue Zhu; Yuhong Geng},
  pdfsubject={Iterative self-labeling for financial market states},
  pdfkeywords={financial time series, market regime classification, iterative self-labeling, structured prediction, label quality evaluation, learnability-discriminability tradeoff}
}
\makeatletter
\renewcommand\footnotesize{\@setfontsize\footnotesize{10}{12}}
\makeatother

\titleformat{\section}{\normalfont\fontsize{14}{17}\selectfont\bfseries}{\thesection}{1em}{}
\titleformat{\subsection}{\normalfont\fontsize{12}{14}\selectfont\bfseries}{\thesubsection}{1em}{}
\titleformat{\paragraph}[runin]{\normalfont\normalsize\bfseries}{}{0pt}{}
\title{\fontsize{14}{17}\selectfont\bfseries DDSL: Iterative Self-Labeling for Financial Market States}
\author{\normalsize Xianbin Gu \qquad Qingyue Zhu \qquad Yuhong Geng\\[4pt]
\normalsize New York University Shanghai, Shanghai, China\\[3pt]
\small \href{mailto:xianbin.gu@nyu.edu}{xianbin.gu@nyu.edu},
\href{mailto:qz2558@nyu.edu}{qz2558@nyu.edu},
\href{mailto:yg3327@nyu.edu}{yg3327@nyu.edu}}
\date{}

\begin{document}
\maketitle
\begin{abstract}
Financial market states lack a unique ground truth and must therefore be
constructed from price data. Conventional labeling methods typically define
states independently of the downstream classifier, leaving unresolved whether
the resulting labels can be recovered from information available at prediction
time. We propose DDSL, an iterative self-labeling framework that couples a
price- and grammar-guided Viterbi labeler with a Transformer classifier through
model feedback. DDSL constructs a nine-state representation that jointly
encodes current and future market trends while retaining structured price
emissions and temporal constraints throughout refinement.

Experiments on SSE50, CSI300, CSI500, and NASDAQ100 show that DDSL increases
held-out joint-state macro F1 by $78\%$--$96\%$ over single-pass
dynamic-programming labeling. An independent GRU classifier that never
participates in the labeling loop reproduces the same trend, while label
entropy and class coverage remain stable, indicating that the gain transfers
across architectures and is not caused by label collapse. We further show
that learnability is controlled by both iteration and the model-feedback
weight $\beta_\text{model}$. Iterative gains diminish as successive label
systems converge, whereas stronger feedback increasingly widens the performance gap between the feedback Transformer and the independent GRU. Increased learnability is accompanied by declining segment-level price-feature separation after an initial improvement. Finally, a
non-anticipative event study provides partial external validation: refinement
strengthens a conditional BEAR$\rightarrow$BULL reversal signal on the three
A-share datasets, while the corresponding effect has the opposite sign on NASDAQ100. These results position DDSL as a structured framework for constructing market-state labels whose learnability
can be explicitly controlled and evaluated.

\end{abstract}


% Publication and license metadata inherited from the ACM manuscript.
\medskip
\noindent{\small Conference version: ICAIF '26.
\href{https://doi.org/10.1145/3858063.3858786}{doi:10.1145/3858063.3858786}.\\
This work is licensed under a
\href{https://creativecommons.org/licenses/by/4.0/}{Creative Commons Attribution 4.0 International License}.}
\medskip

\section{Introduction}
% ==============================================================

Classifying financial time series into bull, bear, and neutral market
states supports applications such as risk management, portfolio
construction, and trading-signal generation. Unlike conventional
supervised classification, however, market-state prediction lacks a
unique, generally accepted ground truth. Supervision must instead be
constructed from price data, and different labeling procedures may produce
different state definitions and boundaries. Discrete states provide a compact representation of price dynamics that can serve as conditioning variables for downstream financial decision models.

Existing approaches, including dynamic-programming
segmentation~\cite{bellman1961approximation} and HMM-based regime
detection~\cite{hamilton1990analysis}, construct market states using a
labeling model specified independently of the downstream classifier. Even
when the labeling procedure itself is estimated iteratively, the classifier
has no mechanism to indicate which state boundaries are recoverable from
the information available at prediction time or which states are
systematically confusable. Consequently, a labeling may describe historical
price behavior while remaining difficult to learn out of sample.

We propose \textbf{DDSL} (Data-Driven Self-Labeling)\footnote{Code is available at \url{https://github.com/hogwild/DDSL}.}, a closed-loop
framework that iteratively couples a structured Viterbi labeler with a
Transformer classifier. Each label jointly represents the trend in a
current window and a future window, producing a nine-state
current--future market-state space with explicit directional
interpretation.

Starting from labels generated without model feedback, DDSL alternates
between training the classifier and incorporating its log-probabilities
into the Viterbi energy. Unlike ordinary pseudo-labeling, the classifier
outputs do not replace the structured labeling objective: price-based
emissions and temporal grammar penalties remain active throughout the
loop. Model feedback can therefore adjust state boundaries toward
configurations that are more recoverable from observable inputs while
retaining the price and temporal structure of the labeling system.

Our contributions are as follows.
\begin{enumerate}
\sloppy
  \item \textbf{DDSL}: We couple a price- and grammar-guided Viterbi
  labeler with a Transformer through model feedback to construct nine joint
  current--future states. Across four datasets, refinement improves label
  learnability, and an independent GRU reproduces the gain, demonstrating
  cross-architecture transfer rather than only the Transformer fitting labels shaped by its own feedback.

  \item \textbf{Characterizing the learnability gain}: Iteration and
  $\beta_\text{model}$ jointly control learnability. At fixed weight, gains
  diminish as the labels converge; increasing the weight improves both classifiers, but an increasing share of the gain accrues to the feedback Transformer. Both routes initially improve and then reduce $\text{FR}_\text{seg}$, exposing the
  cost of increasingly model-aligned labels.

  \item \textbf{External validation}: A non-anticipative event study finds
  that refinement strengthens the BEAR$\rightarrow$BULL reversal signal on
  three A-share datasets but  finds an effect of the opposite sign on NASDAQ100, indicating limited, market-\hspace{0pt}dependent economic content.
\fussy
\end{enumerate}

% ==============================================================
\section{Related Work}
% ==============================================================

\paragraph{Financial time-series labeling.}
Rule-based methods define market states through handcrafted conditions
derived from technical analysis, such as price thresholds, turning-point
detection, or moving-average crossovers~\cite{hanna2018top, zegadlo2022identifying, zakamulin2017market}.
The dominant paradigm for constructing market-state labels applies dynamic
programming or HMM-based segmentation~\cite{bellman1961approximation,
hamilton1990analysis, wu2021optimal} directly to price or return series, with
threshold-based variants defining states by return magnitude over a fixed
lookback window~\cite{kovacevic2023trend}. Triple-barrier labeling~\cite{lopezdeprado2018advances}, the standard construction in financial machine learning, instead labels each observation by which of an upper, lower, or vertical barrier the price path reaches first, treating observations independently of their neighbors.
These approaches construct labels independently of the downstream
classifier and provide no mechanism for incorporating its feedback.
Closest to our work, Ma et al.~\cite{ma2022denoised} generate denoised
financial labels via self-supervised learning and evaluate label quality
through downstream classification and trading performance, but do not
feed model predictions back into the labeling process itself---labels
are produced once, not iteratively refined.
Procacci and Aste~\cite{procacci2019forecasting} define market states via
a sparse precision matrix and forecast state transitions out-of-sample,
but their states are derived from cross-sectional correlation structure
across assets rather than per-instrument price segmentation, and are not
refined through classifier feedback.
To our knowledge, DDSL is the first framework that iteratively feeds
classifier logits back into a price- and grammar-guided Viterbi
labeler for joint current--future market-state construction.
 
\paragraph{Self-training and pseudo-labeling.}
Iterative self-labeling has been widely studied in computer vision and
natural language processing.
Pseudo-labeling~\cite{lee2013pseudo} assigns hard labels from model
predictions and retrains iteratively; FixMatch~\cite{sohn2020fixmatch}
combines pseudo-labeling with consistency regularization for
semi-supervised learning.
In the noisy-label literature, Reed et al.~\cite{reed2014training} propose
bootstrapping, which softens training targets by mixing model predictions
with original labels---a mechanism structurally related to DDSL's energy
interpolation between price signal and model logits.
Tanaka et al.~\cite{tanaka2018joint} jointly optimize labels and model
parameters by treating labels as latent variables, analogous to the
EM-like fixed-point dynamics of DDSL.
In NLP, Yarowsky~\cite{yarowsky1995unsupervised} introduced iterative
self-training for word sense disambiguation, establishing the paradigm
that DDSL applies to the structured financial labeling setting.
 
DDSL differs from these approaches in two fundamental respects.
First, labels are \emph{structurally generated} from price sequences via a grammar-regularized Viterbi decoder~\cite{lafferty2001conditional}, rather than inferred from a fixed set of human-annotated categories.
Second, state-specific price emissions and temporal grammar penalties
remain active throughout refinement, preserving price-based evidence and
temporal coherence rather than replacing the structured labeler with the
model's per-timestep predictions.
 
\paragraph{Label discriminability evaluation.}
Fisher's Linear Discriminant~\cite{fisher1936use} provides a classic
ratio of between-class to within-class variance. We apply this ratio to
segment-level financial features, treating each contiguous same-state
segment rather than each daily observation as the unit of analysis. By
aggregating slope and cumulative return, $\text{FR}_\text{seg}$ is designed
to measure price-feature separation at the granularity of market-state
segments and to respond directly to changes in segment boundaries.

% ==============================================================
\section{Problem Formulation}
\label{sec:formulation}
% ==============================================================

Let $\mathbf{p} = (p_1, \ldots, p_T)$ denote the adjusted daily
closing-price sequence of a single stock, using the market-specific
adjustment conventions described in Section~\ref{sec:experiments}.
The labeling problem is to assign, at each time $t$, a joint market state
$s_t \in \mathcal{S}$ that jointly describes the current-window trend and
the future-window trend.
 
\paragraph{9-state joint labeling.}
A simple 3-class labeling (BEAR / NEUTRAL / BULL) describes the market's
current state but encodes no future-window direction.
Instead, we adopt a \emph{joint} formulation in which each label
simultaneously encodes both the \emph{current-window trend} and the
\emph{future-window trend}.
Each axis takes values in $\{\text{BEAR}, \text{NEUTRAL}, \text{BULL}\}$,
yielding the Cartesian product state space
$\mathcal{S} = \{\text{BEAR}, \text{NEUTRAL}, \text{BULL}\}^2$,
$|\mathcal{S}| = 9$.
States are written in \textsc{current\_to\_future} notation, e.g.,
\texttt{BULL\_TO\_BEAR} denotes a currently bullish regime whose future
window is labeled bearish.
 
This joint design has a consequence that is central to the entire method:
because each label encodes \emph{both} the present and the near-future,
adjacent labels are not independent.
Specifically, if the label at time $t$ states that the future trend is
$s_{f,t}$, then the label at time $t{+}1$ should have a current trend
$s_{c,t+1}$ consistent with $s_{f,t}$---since the future window of $t$ and
the current window of $t{+}1$ overlap substantially.
This consistency requirement defines a \emph{label grammar}:
\[
  \text{Grammar rule: } s_{f,t} \approx s_{c,t+1} \quad \forall\, t.
\]
The grammar is not a hard constraint (market regimes can transition
abruptly), but its soft enforcement via $E_\text{grammar}$ in the Viterbi
objective---described in Section~\ref{sec:dp}---is what distinguishes our
labeling from simple independent per-timestep classification.
$E_\text{grammar}$ unifies three structural penalties: continuity of the
current-trend axis, continuity of the future-trend axis, and the
cross-timestep consistency rule $s_{f,t} \approx s_{c,t+1}$.
The grammar acts as a temporal regularizer: every inconsistent
transition incurs a penalty independent of $\beta_\text{model}$. It does
not mathematically rule out a grammar-consistent collapsed sequence;
rather, together with the state-specific price emissions, it constrains
how model feedback can reshape the label sequence. We therefore assess
state preservation empirically in Section~\ref{sec:learnability}.
 
\paragraph{Learnability metric.}
How easily a classifier can learn the current label set is measured by
\emph{joint-state macro F1} ($\text{F1}_\text{macro}$): the unweighted
average of per-class F1 scores across all 9 states on the held-out test
split.
Macro averaging prevents the neutral-dominant class from masking
minority-state performance.
A higher $\text{F1}_\text{macro}$ indicates that the label set is more
learnable---its class boundaries are better aligned with patterns
extractable from the token sequence. Unless otherwise noted,
$\text{F1}_\text{macro}$ is evaluated with the feedback Transformer
(Section~\ref{sec:transformer}); Section~\ref{sec:learnability} shows this
learnability gain is not specific to that architecture.
 
\paragraph{Discriminability metric.}
Separation of the nine joint states in segment-level price-feature space
is measured by $\text{FR}_\text{seg}$, the equal-weight macro-average of
per-feature Fisher Ratios over $\mathcal{F}_\text{seg}$, comprising
\texttt{seg\_slope} and \texttt{dC\_sum}.
Section~\ref{sec:analysis} gives the operational definition. Higher
$\text{FR}_\text{seg}$ indicates better-separated, more compact state
regions in this feature space.
 
\paragraph{Iteration notation.}
We index iterations as $\text{iter}_k$, $k = 0, 1, \ldots, K_\text{iter}$,
where $k{=}0$ denotes the pure-DP baseline with no model feedback, and
$k = 1, \ldots, K_\text{iter}$ denote the DDSL iterations.
The labels at $\text{iter}_k$ ($k \geq 1$) depend on the classifier
logits exported at $\text{iter}_{k-1}$.
At every iteration and for every data split, Viterbi decoding is performed
over an expanding historical prefix ending at the corresponding split
boundary, so that val- and test-split labels are conditioned on all
available history up to that point.

% ==============================================================
\section{Method}
\label{sec:method}

% --------------------------------------------------------------
\subsection{K/Q/G/M Tokenization}
\label{sec:tokenization}
% --------------------------------------------------------------

Raw OHLCV data for each stock symbol is converted into a sequence of
discrete tokens before labeling or model training. We use the K/Q/G/M
tokenization scheme with a shared token vocabulary.

Each daily bar is described by four components:
\begin{itemize}
  \item \textbf{K} (geometry): candlestick geometry is represented by four
  normalized OHLC-derived features (body-to-range ratio, close-location
  value, upper-wick ratio, and lower-wick ratio). These feature vectors are
  partitioned into $K_n$ clusters using $k$-means, and the corresponding
  cluster index is used as the geometry token.
  \item \textbf{Q} (daily return): log-return
  $\Delta C=\log(C_t/C_{t-1})$, quantized into $Q_n$ bins via pre-fitted
  bin edges.
  \item \textbf{G} (gap): overnight gap
  $\textit{gap}=\log(O_t/C_{t-1})$, quantized into $G_n$ bins.
  \item \textbf{M} (turnover): daily turnover, quantized into $M_n$
  bins using per-symbol bin edges.
\end{itemize}

The K, Q, and G mappings are fitted on the training split and shared across
symbols within each market. For M, all symbols use the same $M_n$ tokens,
but the bin edges are fitted separately from each symbol's training
history. All mappings are fixed across validation, test, and DDSL
iterations. The Transformer embeds $K$, $Q$, $G$, and $M$ separately
(Section~\ref{sec:transformer}).

% --------------------------------------------------------------
\subsection{DP Baseline Labeling (\texorpdfstring{iter$_0$}{iter0})}
\label{sec:dp}
% --------------------------------------------------------------
 
The iter$_0$ labels are produced by a pure dynamic programming labeler with no model feedback.
For each stock, Viterbi decoding is performed over the EWMA-smoothed price sequence following the expanding-prefix convention defined in Section~\ref{sec:formulation}.
 
The emission energy for assigning state $s \in \mathcal{S}$ to time $t$ is:
\[
  E_\text{price}(t, s) = \lambda_c \cdot \phi(t, s_c;\, \text{current}) + \lambda_f \cdot \phi(t, s_f;\, \text{future}),
\]
where $s_c$ and $s_f$ are the current and future trend components of $s$, and $\phi(t, \tau;\cdot)$ scores the agreement between a price window's statistics and a trend label $\tau \in \{\text{BEAR}, \text{NEUTRAL}, \text{BULL}\}$.
Three statistics are computed on the EWMA-smoothed series $\tilde{p}$ for each window: normalized OLS slope $\alpha = \frac{\sum_i (i-\bar{i})(\tilde{p}_i - \bar{\tilde{p}})}{\sum_i(i-\bar{i})^2} \cdot \frac{1}{|\bar{\tilde{p}}|}$; flip rate $\rho$ (fraction of consecutive return-sign changes, measuring choppiness); and positivity ratio $\pi$ (fraction of positive non-zero returns).
These combine into trend scores:
\begin{equation*}
\begin{array}{l}
  \phi(\tau{=}\text{BULL})    = w_\alpha(\alpha - \tau_0) - w_\rho \rho + w_\pi(\pi - 0.5), \\[4pt]
  \phi(\tau{=}\text{BEAR})    = w_\alpha(-\alpha - \tau_0) - w_\rho \rho + w_\pi(0.5 - \pi), \\[4pt]
  \phi(\tau{=}\text{NEUTRAL}) = -w_\alpha|\alpha| + w_{\rho,n}\rho - w_{\pi,n}|\pi - 0.5|,
\end{array}
\end{equation*}
where $\tau_0$ is a slope dead-band threshold and $w_\cdot$ are fixed weights.
Current-window statistics use the symmetric window $[\,t{-}L_c,\, t{+}L_c\,]$; future-window statistics use $[\,t{+}1,\, t{+}L_f\,]$. Both windows are clipped to the available decoded prefix so that no prices beyond the corresponding split cutoff are used.
For the final two observations, fewer than two future prices remain,
so $E_\text{price}$ is identical across states; labels then depend on
$E_\text{grammar}$ and, for $k \geq 1$, model feedback.
These boundary labels are retained without additional endpoint trimming.
The grammar energy encodes all structural constraints from the 9-state design:
\begin{align}
  E_\text{grammar}(s_t, s_{t+1}) =\;&\psi_\lambda(s_{c,t},\, s_{c,t+1}) \notag \\
                                    +\;&\psi_\lambda(s_{f,t},\, s_{f,t+1}) \notag \\
                                    +\;&\psi_\mu(s_{f,t},\, s_{c,t+1}),
  \label{eq:egrammar}
\end{align}
where $\psi_\lambda(\cdot,\cdot)$ and $\psi_\mu(\cdot,\cdot)$ share the
same non-positive form---zero for identical trends, a neutral penalty if
one trend is neutral, an opposite penalty for opposite trends---but use
different weights: $\psi_\lambda$ uses $\lambda_\text{neutral}$ and
$\lambda_\text{opp}$, whereas $\psi_\mu$ uses the weaker
$\mu_\text{neutral} \leq \lambda_\text{neutral}$ and $\mu_\text{opp} \leq
\lambda_\text{opp}$ to permit genuine regime transitions.
The first two terms penalize abrupt changes along the current and future axes, respectively; the third enforces $s_{f,t} \approx s_{c,t+1}$, which is unique to the 9-state formulation and has no analog in 3-state schemes.
$E_\text{grammar}$ remains active in both the DP baseline and all DDSL iterations.
 
The total DP objective is:
\[
  \mathbf{s}^* = \arg\max_{\mathbf{s}} \sum_{t=1}^{T} E_\text{price}(t, s_t) + \sum_{t=1}^{T-1} E_\text{grammar}(s_t, s_{t+1}).
\]
Viterbi decoding finds the exact global optimum in $O(T \cdot |\mathcal{S}|^2)$ time~\cite{viterbi1967error}.

% --------------------------------------------------------------
\subsection{DDSL Iterative Labeling (\texorpdfstring{iter$_k$, $k \geq 1$}{iter k, k >= 1})}
\label{sec:ddsl}
% --------------------------------------------------------------
 
At iteration $k \geq 1$, the DDSL labeler augments the DP energy with a model feedback term.
Let $\ell_{k-1}(t, s) = \log\text{-softmax}_s(z_{k-1}(t,\cdot)/\tau_\text{soft})$ denote the temperature-scaled log-probability derived from the iter$_{k-1}$ classifier's raw logits $z_{k-1}(t,s)$.
The DDSL energy is:
\begin{equation}
  E_\text{DDSL}(t, s_t, s_{t+1}) = E_\text{price}(t, s_t) + E_\text{grammar}(s_t, s_{t+1}) + \beta_\text{model} \cdot \ell_{k-1}(t, s_t).
  \label{eq:ddsl}
\end{equation}
The three terms have distinct roles: $E_\text{price}$ supplies price-signal evidence, $E_\text{grammar}$ regularizes temporal structure, and $\beta_\text{model} \cdot \ell_{k-1}$ injects model feedback.
 
Three properties of Eq.~\eqref{eq:ddsl} matter.
(1) Normalization does not affect the decoded path. We use temperature-scaled
log-probabilities ($\tau_\text{soft}{=}1$). For a fixed-length sequence,
log-softmax normalization contributes the same additive constant to
every candidate path. Decoding is therefore equivalent to using
temperature-scaled logits with the same feedback weight, giving an
effective coefficient $\beta_\text{model}/\tau_\text{soft}$ on the raw logits.
(2) Additive energy, not label replacement. Model feedback
reshapes the energy landscape while retaining price, grammar, and
structured decoding.
(3) $\beta_\text{model}{=}0$ recovers pure DP, making iter$_0$
a direct reference for the feedback ablation.
 
\begin{figure}[t]
  \centering
  \includegraphics[width=\linewidth]{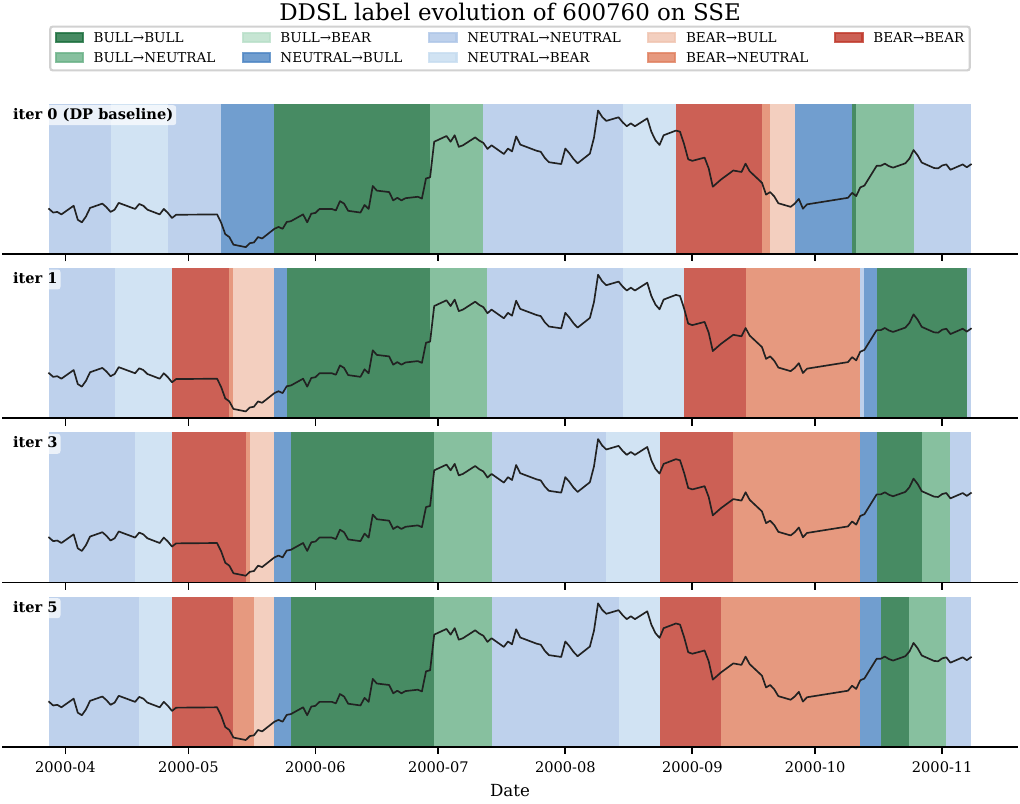}
  \caption{
    DDSL label evolution at selected iterations for stock 600760 (Shanghai Stock Exchange, 2000/03/28--2000/11/08). Segment boundaries shift under model feedback, with progressively smaller changes at later iterations.
  }
% Figure description: Price chart of one Shanghai-listed stock over eight months in 2000, with the nine-state labels assigned at several DDSL iterations; segment boundaries move between early iterations and change little in later ones.
  \label{fig:ddsl-label-evolution-example}
\end{figure}

Algorithm~\ref{alg:ddsl} summarizes the full loop; Figure~\ref{fig:ddsl-label-evolution-example} shows an example of label evolution across iterations.
 
\begin{algorithm}[t]
\caption{DDSL Iterative Self-Labeling}
\label{alg:ddsl}
\begin{algorithmic}
\REQUIRE Price sequences $\{\mathbf{p}^{(i)}\}$, tokens $\{(\mathbf{K}^{(i)}, \mathbf{Q}^{(i)}, \mathbf{G}^{(i)}, \mathbf{M}^{(i)})\}$, iterations $K_\text{iter}$, weight $\beta_\text{model}$
\STATE $\mathbf{L}^0 \leftarrow \text{DPLabeler}(\{\mathbf{p}^{(i)}\})$ \COMMENT{iter$_0$: pure DP}
\STATE $\theta_0 \leftarrow \text{TrainTransformer}(\{(\mathbf{K}^{(i)}, \mathbf{Q}^{(i)}, \mathbf{G}^{(i)}, \mathbf{M}^{(i)}), \mathbf{L}^0\})$
\FOR{$k = 1$ \TO $K_\text{iter}$}
  \STATE $\mathbf{Z}^{k-1} \leftarrow \text{ExportLogits}(\theta_{k-1}, \{(\mathbf{K}^{(i)}, \mathbf{Q}^{(i)}, \mathbf{G}^{(i)}, \mathbf{M}^{(i)})\})$
  \STATE $\mathbf{L}^k \leftarrow \text{ViterbiDecode}\bigl(E_\text{price} + E_\text{grammar} + \beta_\text{model} \cdot \log\text{-softmax}(\mathbf{Z}^{k-1}/\tau_\text{soft})\bigr)$
  \STATE $\theta_k \leftarrow \text{TrainTransformer}(\{(\mathbf{K}^{(i)}, \mathbf{Q}^{(i)}, \mathbf{G}^{(i)}, \mathbf{M}^{(i)}), \mathbf{L}^k\})$
\ENDFOR
\RETURN $\mathbf{L}^{K_\text{iter}}$
\end{algorithmic}
\end{algorithm}
 
% --------------------------------------------------------------
\subsection{Transformer Classifier}
\label{sec:transformer}
% --------------------------------------------------------------
 
At each DDSL iteration, a Transformer encoder~\cite{vaswani2017attention} is trained on the current label set and used to export per-token logits for the next relabeling round.
 
\paragraph{Input representation.}
The model receives a sliding window of $T_\text{seq}$ consecutive tokens.
Each token at position $i$ is represented by four separate embedding lookups concatenated and projected into a shared $d$-dimensional space:
\[
  \mathbf{h}_i = \mathbf{W}_\text{proj}\bigl[\mathbf{E}_K[K_i] \,\Vert\, \mathbf{E}_Q[Q_i] \,\Vert\, \mathbf{E}_G[G_i] \,\Vert\, \mathbf{E}_M[M_i]\bigr] \in \mathbb{R}^{d},
\]
where each $\mathbf{E}_\cdot$ is a learned embedding table and $\mathbf{W}_\text{proj} \in \mathbb{R}^{d \times 4d}$ fuses the four factors.
A learned positional embedding is added to encode sequence order.
 
\paragraph{Encoder and pooling.}
The token sequence is passed through a standard Transformer encoder.
The \emph{last}-token output is used as the sequence representation (\texttt{last}-token pooling): since the task is to predict the joint state $s_t$ at the window endpoint, this provides the most direct signal about the current timestep while retaining full historical context.
 
\paragraph{Output and training.}
A linear head projects the pooled representation to $|\mathcal{S}|{=}9$ logits.
The model is trained with class-weighted focal loss~\cite{lin2017focal} (inverse-square-root class weights, normalized to unit mean) to mitigate joint-state class imbalance.
At logit export, raw pre-softmax logits for all 9 states are written to disk for every token position across all splits (train, val, test), providing uniform model feedback to the DDSL labeler across the full time series.

% ==============================================================
\subsection{Measuring Label Discriminability}
\label{sec:analysis}
% ==============================================================
 
Since ground truth does not exist, we assess label discriminability through a segment-level Fisher Ratio that measures how well the 9 joint states are geometrically separated in segment-level price-feature space.
 
\paragraph{Segment-level Fisher Ratio ($\text{FR}_\text{seg}$).}
Each contiguous same-state segment is summarized by two features:
\begin{itemize}
  \item \textbf{seg\_slope}: OLS slope on the EWMA-smoothed price series within the segment, normalized by mean segment price. Positive for bull segments, negative for bear, near-zero for neutral.
  \item \textbf{dC\_sum}: sum of daily log-returns within the segment, capturing both return intensity and segment duration.
\end{itemize}
For each feature $f \in \mathcal{F}_\text{seg} = \{\texttt{seg\_slope},\, \texttt{dC\_sum}\}$, the per-feature Fisher Ratio is:
\[
  \text{FR}(f) = \frac{\displaystyle\frac{1}{|\mathcal{S}|}\sum_{s \in \mathcal{S}} (\mu_s^f - \mu^f)^2}
                      {\displaystyle\frac{1}{|\mathcal{S}|}\sum_{s \in \mathcal{S}} \sigma_s^{f,2} + \epsilon},
\]
where $\mu_s^f$ and $\sigma_s^{f,2}$ are the mean and variance of feature $f$ over all segments of state $s$, $\mu^f$ is the equal-weight global mean across states, and $\epsilon$ prevents division by zero.
Equal-weight averaging across all $|\mathcal{S}|{=}9$ states prevents majority-state dominance.
The scalar $\text{FR}_\text{seg}$ is the macro-average of $\text{FR}(f)$ over $\mathcal{F}_\text{seg}$, and is computed on the whole dataset.
Its evolution across iterations directly tracks the Learnability--Discriminability tradeoff described in Section~\ref{sec:learnability}.
 
% ==============================================================
\section{Experiments}
\label{sec:experiments}
% ==============================================================
 
% --------------------------------------------------------------
\subsection{Experimental Setup}
\label{sec:setup}
% --------------------------------------------------------------
 
\paragraph{Datasets and splits.}
We evaluate on four datasets covering two markets.
The three Chinese datasets are drawn from the constituent stocks of major
A-share equity indices: \textbf{SSE50} (Shanghai Stock Exchange 50,
large-cap, 40/50 stocks retained), \textbf{CSI300} (China Securities Index
300, broad market, 248/300), and \textbf{CSI500} (CSI 500, mid- and small-cap,
427/500). The fourth dataset uses constituent stocks of the
\textbf{NASDAQ100} (91/101 US stocks retained; the index includes 101
securities due to dual share classes). 

These datasets were selected to cover both Chinese and US equity markets
while spanning a broad range of market capitalizations. Because price dynamics can vary with firm size, large-cap stocks
(SSE50 and NASDAQ100), broad-market stocks (CSI300), and mid- and
small-cap stocks (CSI500) provide a diverse test bed for evaluating the
robustness of DDSL across different market structures.

Daily A-share data were downloaded using AKShare, and NASDAQ100 data
were downloaded from Yahoo Finance. All constituent lists are fixed as
of December~31, 2025. The observation period is January~1,
2000--December~31, 2025; later-listed stocks begin on their first available
trading date. Chinese prices use backward adjustment (HFQ), incorporating
dividends and other corporate actions, whereas NASDAQ prices use fully
adjusted series that incorporate both stock splits and cash dividends.
% Stocks with non-positive adjusted prices or insufficient history before
% the training boundary are excluded via a per-dataset blacklist.
Rows with a missing open, high, low, or close are dropped at download;
no values are imputed.

A per-dataset blacklist excludes a stock from all splits if any adjusted
open, high, low, or close value is non-positive, or if fewer than 60 input
records remain on or before December~31, 2018. This history threshold
is evaluated before tokenization. The same blacklist is used
for codebook fitting, labeling, and dataset construction. During
tokenization, rows with missing candlestick-geometry, daily-return,
overnight-gap, or turnover features are dropped, again without
imputation. Because index membership is fixed to a December~2025 snapshot, the results
describe the retained constituents rather than historical point-in-time
index universes; we therefore make no index- or portfolio-level return
claims. Splits are train ($\leq$2018), validation (2019--2022), and test
(2023--2025). Labels are generated via expanding historical prefixes
ending at each boundary, and only the corresponding target-period labels
are retained. Fisher Ratio metrics are computed over the entire labeled
period.
 
\paragraph{Hyperparameters.}
Codebook sizes are $K_n{=}16$, $Q_n{=}8$, $G_n{=}8$, and $M_n{=}4$.
The classifier is a 6-layer Transformer encoder with
$d_\text{model}{=}64$, 4 heads, feedforward dimension~128, dropout~0.3,
sequence length~60, and last-token pooling. It is trained for 60~epochs
with AdamW ($\text{lr}{=}3{\times}10^{-4}$, weight decay $10^{-4}$, batch size 2048, gradient clipping at 1.0, ReduceLROnPlateau) and
class-weighted focal loss ($\gamma{=}2$); the best validation macro-F1
checkpoint is used for logit export. Main-configuration labeling
parameters are listed in Table~\ref{tab:energy_hparams}; 
$\beta_\text{model}$ sensitivity is examined in
Section~\ref{sec:ablation}. $\beta_\text{model}{=}0.3$ was fixed before the ablation
experiments and is shared across all datasets.

\begin{table}[t]
\caption{Main-configuration labeling hyperparameters, shared across
all datasets, seeds, and iterations unless varied in the
$\beta_\text{model}$ ablation. $L_c$ is the half-width of the symmetric
current window.}
\label{tab:energy_hparams}
\centering
\small
\setlength{\tabcolsep}{4pt}
\begin{tabular}{@{}lclc@{}}
\toprule
\textbf{Parameter} & \textbf{Value}
& \textbf{Parameter} & \textbf{Value} \\
\midrule
\multicolumn{4}{c}{\textit{Smoothing and windows}} \\
EWMA span & $10$ & $L_c$ & $10$ \\
$L_f$ & $20$ & $\tau_0$ & $5{\times}10^{-4}$ \\
\midrule
\multicolumn{4}{c}{\textit{Price-emission scoring}} \\
$\lambda_c$ & $1.0$ & $\lambda_f$ & $1.2$ \\
$w_\alpha$ & $1.0$ & $w_\rho$ & $0.8$ \\
$w_\pi$ & $0.6$ & $w_{\rho,n}$ & $0.4$ \\
$w_{\pi,n}$ & $0.6$ & & \\
\midrule
\multicolumn{4}{c}{\textit{Grammar penalties}} \\
$\lambda_\text{neutral}$ & $0.25$ & $\lambda_\text{opp}$ & $1.00$ \\
$\mu_\text{neutral}$ & $0.05$ & $\mu_\text{opp}$ & $0.10$ \\
\midrule
\multicolumn{4}{c}{\textit{Model feedback}} \\
$\beta_\text{model}$ & $0.3$ & $\tau_\text{soft}$ & $1.0$ \\
\bottomrule
\end{tabular}
\end{table}

\paragraph{GRU learnability probe.}
At each iteration, we train an independent two-layer GRU
(hidden size~64, dropout~0.3) using the same K/Q/G/M front end, training
schedule, loss, and class weights as the Transformer. Its logits never
enter the labeler, making it a read-only cross-architecture probe used in
Sections~\ref{sec:learnability} and~\ref{sec:ablation}.
 
\paragraph{Seeds and iterations.}
All experiments use seeds $\{41,42,43\}$; seed-dependent results are
reported as mean~$\pm$~std. We run iter$_0$ (DP baseline) through
iter$_5$ (five DDSL iterations), by which macro-F1 increases have
substantially diminished across all datasets.
At each iteration, both the Transformer and the GRU are reinitialized and trained from scratch, without carrying over weights from the previous iteration.

% --------------------------------------------------------------
\subsection{Label Learnability}
\label{sec:learnability}
% --------------------------------------------------------------
\begin{table}[t]
\caption{Summary of label learnability and its cost. F1 and
$\text{FR}_{\mathrm{seg}}$ entries are means over three seeds.
$\Delta H$ is the relative change in normalized label entropy from
iter$_0$ to iter$_5$.}
\label{tab:learnability_summary}
\centering
\small
\setlength{\tabcolsep}{3pt}
\begin{tabular}{lcccc}
\toprule
\textbf{Dataset}
& \textbf{F1(Tr.) $0{\to}5$}
& \textbf{F1(GRU) $0{\to}5$}
& $\boldsymbol{\Delta H}$
& $\mathbf{FR}_{\mathrm{seg}}\ \mathbf{0{\to}1{\to}5}$ \\
\midrule
SSE50
& $.246{\to}.482$
& $.245{\to}.460$
& $+0.83\%$
& $.668{\to}.772{\to}.647$ \\

CSI300
& $.247{\to}.458$
& $.253{\to}.429$
& $+0.01\%$
& $.791{\to}.933{\to}.730$ \\

CSI500
& $.246{\to}.460$
& $.245{\to}.441$
& $+1.13\%$
& $.744{\to}.868{\to}.648$ \\

NASDAQ100
& $.231{\to}.411$
& $.240{\to}.385$
& $-0.78\%$
& $.727{\to}.840{\to}.662$ \\
\bottomrule
\end{tabular}
\end{table}

\begin{figure}[t]
  \centering
  \includegraphics[width=\linewidth]{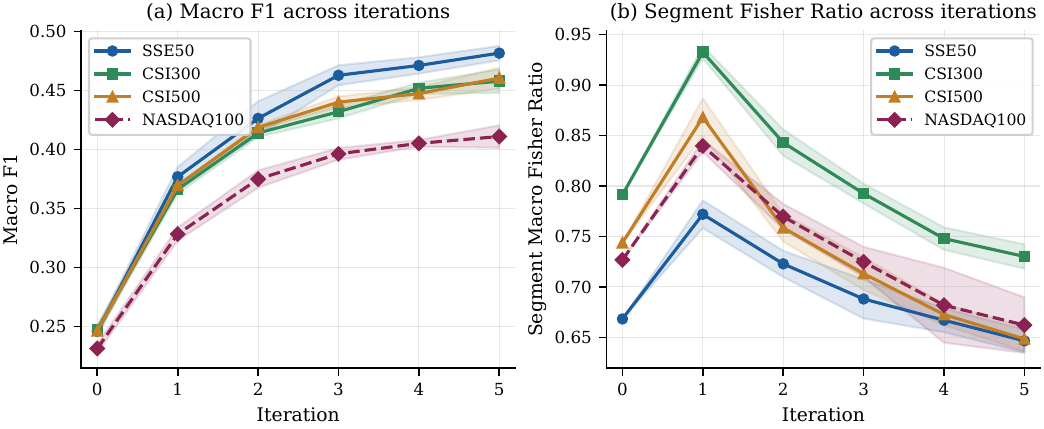}
\caption{Full iteration trajectories underlying
Table~\ref{tab:learnability_summary}. Transformer joint-state macro F1
increases across all four datasets (a), whereas
$\text{FR}_\text{seg}$ peaks at iter$_1$ and then declines (b).
Lines show means over three seeds; shaded bands indicate
$\pm1$ standard deviation.}
% Figure description: Two line panels over iterations 0 to 5 for four datasets: Transformer macro F1 rises in every dataset, while the segment-level Fisher Ratio peaks at iteration 1 and then declines; shaded bands show one standard deviation across seeds.
\label{fig:co_evolution}
\end{figure}

\paragraph{Interpretation.}
Because financial market states have no fixed ground truth, macro F1 at
iteration $k$ is computed against the corresponding labels
$\mathbf{L}^{k}_{\mathrm{test}}$. Comparisons across iterations therefore
measure the \emph{recoverability} of each label system from observable
inputs, rather than performance on an invariant classification task. This
is an important property because labels intended for prediction must be
learnable from information available at prediction time. We additionally
test whether an apparent gain could instead arise from Transformer
self-confirmation or label-distribution collapse.

\paragraph{DDSL increases label learnability across markets and
architectures.}
As summarized in Table~\ref{tab:learnability_summary}, Transformer macro F1, i.e., F1(Tr.), increases from $.231$--$.247$ under the single-pass DP labels
(iter$_0$) to $.411$--$.482$ at iter$_5$, corresponding to relative gains
of $77.9\%$--$95.9\%$. The increase is monotone in all twelve
dataset--seed runs. Price emissions and grammar penalties remain active
throughout these iterations, so the result is obtained within the structured
DDSL objective rather than by directly copying model predictions. Figure~\ref{fig:co_evolution} shows the full iteration trajectories of the Transformer behind the summary statistics in Table~\ref{tab:learnability_summary}.

The same pattern holds for the independent GRU probe, whose logits never
enter the labeling loop. GRU macro F1 is positively associated with
iteration in all twelve dataset--seed runs, with a strictly monotone
increase in eleven. The improvement therefore transfers to a classifier
that did not generate the feedback, supporting the interpretation that
refinement increases the learnability of the label system rather than
only the Transformer's fit to its own outputs. 

\paragraph{The gain is not caused by label collapse.}
The nine-state distribution remains stable during refinement. From iter$_0$
to iter$_5$, normalized Shannon entropy changes by only $-0.78\%$ to
$+1.13\%$ across datasets, no state disappears, and the smallest state share
over all datasets, iterations, and seeds remains at least $0.61\%$.
These changes are negligible relative to the $78\%$--$96\%$ macro-F1
increase. The improvement therefore cannot be explained by labels
concentrating into fewer classes or by the classification task collapsing
to a simpler state space.

\paragraph{The gain has a cost.}
Segment-level price-feature separation initially improves at iter$_1$ but
then declines monotonically through iter$_5$ on every dataset and seed.
Over iter$_1$--iter$_5$, Spearman's $\rho$ between Transformer macro F1
and $\text{FR}_{\mathrm{seg}}$ is $-1.00$ in all twelve runs, indicating a
consistent inverse ordering during convergence. Relative to the DP baseline,
iter$_5$ sacrifices $3\%$--$13\%$ of $\text{FR}_{\mathrm{seg}}$ for the
learnability gains above; relative to the iter$_1$ peak, the decline is
$16\%$--$25\%$. Because $\text{FR}_{\mathrm{seg}}$ is computed directly
from labels and price features, this cost is architecture-independent.
It should nevertheless be interpreted as one geometric property of the
labels, not as a complete measure of their economic value.

Overall, DDSL produces labels that are increasingly recoverable across
markets and classifier architectures without collapsing the state
distribution, but continued refinement trades away some segment-level
price separation. Sections~\ref{sec:saturation} and~\ref{sec:ablation}
next examine how iteration and $\beta_{\text{model}}$, respectively,
control this gain.

% --------------------------------------------------------------
\subsection{Saturation of the Learnability Gain}
\label{sec:saturation}
% --------------------------------------------------------------

We now examine how much additional improvement is obtained
from continuing the loop while holding
$\beta_\text{model}{=}0.3$ fixed.

\paragraph{The gain is front-loaded.}
As shown in Figure~\ref{fig:co_evolution}, most of the macro-F1 improvement
occurs during the first two feedback rounds. Relative to the total
iter$_0$$\to$iter$_5$ gain, iter$_1$ contributes
$53.9\%$--$57.9\%$, and the first two iterations together contribute
$76.3\%$--$80.8\%$ across the four datasets. Subsequent increments become
progressively smaller. By iter$_4$$\to$iter$_5$, the mean gain is only
$.006$--$.013$, comparable to the across-seed standard deviation on each
dataset. We therefore interpret the trajectories as approaching a plateau,
without claiming that the marginal gain has become exactly zero.

\paragraph{Successive label systems converge.}
The diminishing gain coincides with stabilization of the generated labels.
Table~\ref{tab:convergence} reports the joint-state change rate and Cohen's
$\kappa$ between adjacent iterations. Across all datasets, the proportion
of changed labels falls from approximately $20\%$--$22\%$ after the first
feedback round to $3\%$--$5\%$ between iter$_4$ and iter$_5$, while
$\kappa$ rises from approximately $.73$--$.75$ to $.94$--$.96$.
Thus, later classifiers provide feedback to an increasingly stable label
system rather than repeatedly reorganizing it. Once successive iterations
agree on approximately $95\%$--$97\%$ of test timesteps, little label mass remains
available for further refinement, explaining the diminishing improvement in
recoverability. This behavior is consistent with the labeler--classifier
loop approaching an empirical fixed point.

\paragraph{Scope of the saturation result.}
The observed saturation applies to repeated iteration under a fixed energy
balance. It is not an absolute upper bound on macro F1: changing
$\beta_\text{model}$ alters the relative influence of model feedback,
price emissions, and grammar constraints, and can move the system to a
different operating point. The next section isolates this second factor.
Accordingly, iteration determines how far DDSL progresses toward the fixed
point associated with a given $\beta_\text{model}$, whereas
$\beta_\text{model}$ determines the fixed-point regime being approached.

\begin{table}[t]
\caption{Convergence between adjacent DDSL iterations on the test split,
averaged over three seeds. Each entry reports joint-state change rate
(CR) followed by Cohen's $\kappa$.}
\label{tab:convergence}
\centering
\small
\setlength{\tabcolsep}{4pt}
\begin{tabular}{lcccc}
\toprule
\textbf{Transition}
& \textbf{SSE50}
& \textbf{CSI300}
& \textbf{CSI500}
& \textbf{NASDAQ100} \\
\midrule
$0{\to}1$ & $.209/.741$ & $.214/.735$ & $.220/.729$ & $.198/.749$ \\
$1{\to}2$ & $.096/.881$ & $.099/.877$ & $.100/.877$ & $.088/.887$ \\
$2{\to}3$ & $.058/.929$ & $.077/.904$ & $.066/.919$ & $.060/.924$ \\
$3{\to}4$ & $.047/.942$ & $.058/.929$ & $.057/.930$ & $.039/.951$ \\
$4{\to}5$ & $.040/.951$ & $.050/.938$ & $.050/.939$ & $.032/.960$ \\
\bottomrule
\end{tabular}
\end{table}

% --------------------------------------------------------------
\subsection{Effect of Model-Feedback Strength}
\label{sec:ablation}
% --------------------------------------------------------------
\begin{table}[t]
\caption{Effect of $\beta_\text{model}$ on Transformer macro F1, GRU macro
F1, and segment-level Fisher Ratio at iter$_5$, means over three seeds.
Seed standard deviations do not exceed $.021$ for F1 and $.040$ for
$\text{FR}_\mathrm{seg}$.}
\label{tab:beta_ablation}
\centering
\small
\begin{tabular}{lcccccc}
\toprule
& \multicolumn{3}{c}{\textbf{SSE50}} & \multicolumn{3}{c}{\textbf{NASDAQ100}} \\
\cmidrule(lr){2-4}\cmidrule(lr){5-7}
$\boldsymbol{\beta_\text{model}}$
& \textbf{F1(Tr.)} & \textbf{F1(GRU)} & $\mathbf{FR}_\mathrm{seg}$
& \textbf{F1(Tr.)} & \textbf{F1(GRU)} & $\mathbf{FR}_\mathrm{seg}$ \\
\midrule
$0$    & $.246$ & $.245$ & $.668$ & $.231$ & $.240$ & $.727$ \\
$0.1$  & $.358$ & $.348$ & $.774$ & $.317$ & $.313$ & $.795$ \\
$0.2$  & $.418$ & $.404$ & $.693$ & $.367$ & $.352$ & $.699$ \\
$0.3$  & $.482$ & $.460$ & $.647$ & $.411$ & $.385$ & $.662$ \\
$0.4$  & $.580$ & $.528$ & $.641$ & $.463$ & $.431$ & $.655$ \\
$0.5$  & $.681$ & $.622$ & $.628$ & $.564$ & $.508$ & $.659$ \\
$0.6$  & $.757$ & $.686$ & $.563$ & $.645$ & $.585$ & $.617$ \\
$0.7$  & $.816$ & $.737$ & $.527$ & $.731$ & $.659$ & $.580$ \\
$0.8$  & $.851$ & $.758$ & $.500$ & $.780$ & $.696$ & $.542$ \\
\midrule
Pseudo & $.935$ & $.764$ & $.454$ & $.922$ & $.743$ & $.457$ \\
\bottomrule
\end{tabular}
\end{table}
We now vary $\beta_\text{model}$ to examine how the strength of model
feedback changes the label system. The ablation is conducted at iter$_5$ on
SSE50 and NASDAQ100. The $\beta_\text{model}{=}0$ condition is the pure-DP
reference, which is invariant to iteration. Pseudo-labeling assigns each
timestep the Transformer argmax and removes the price-emission, grammar, and
Viterbi components of DDSL, so it is a structural contrast rather than
another finite point on the DDSL trajectory.
 
\paragraph{Stronger feedback increases learnability.}
Table~\ref{tab:beta_ablation} shows that introducing a small amount of model
feedback improves both learnability and price-feature separation. At
$\beta_\text{model}{=}0.1$, Transformer macro F1 rises from $.246$ to $.358$
while $\text{FR}_\text{seg}$ rises from $.668$ to $.774$; NASDAQ100 moves the
same way, from $.231$ to $.317$ and from $.727$ to $.795$. Beyond this
initial correction, increasing $\beta_\text{model}$ raises macro F1
strictly monotonically for both the Transformer and the independent GRU
probe, on every seed of both datasets, while lowering
$\text{FR}_\text{seg}$ overall. Thus $\beta_\text{model}$ controls the
balance between label recoverability and segment-level price separation:
stronger feedback makes the labels easier to predict, but increasingly
shifts the Viterbi energy toward the model term and away from the price and
grammar terms.

\paragraph{The gain becomes increasingly architecture-specific.}
Under pure DP, the two architectures perform similarly, with
Transformer--GRU gaps of $.001$ on SSE50 and $-.009$ on NASDAQ100.
The gap then widens with feedback strength: it reaches $.010$ and $.004$
at $\beta_\text{model}{=}0.1$, $.093$ and $.084$ at $0.8$, and $.171$
and $.179$ under pseudo-labeling on SSE50 and NASDAQ100, respectively.
From $\beta_\text{model}{=}0.8$ to pseudo-labeling, Transformer macro F1
increases by $.084$ on SSE50 and $.142$ on NASDAQ100, whereas GRU macro
F1 increases by only $.006$ and $.047$. Thus, stronger feedback continues
to improve cross-architecture recoverability, but an increasing share of
the gain accrues to the Transformer whose outputs shape the labels.

Pseudo-labeling produces the highest Transformer macro F1 and the lowest
$\text{FR}_\text{seg}$ on both datasets, at nearly identical coordinates
($.935,.454$) and ($.922,.457$). DDSL is therefore not designed merely
to maximize the feedback Transformer's F1, but to improve recoverability
while retaining price and temporal structure.
 
\begin{figure}[t]
  \centering
  \includegraphics[width=0.95\columnwidth]{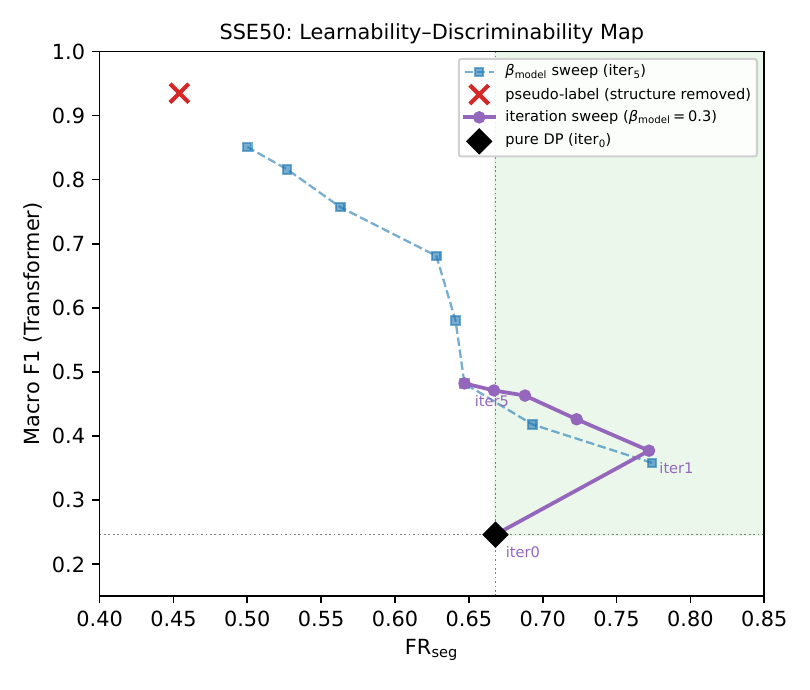}
  \caption{Learnability--discriminability map on SSE50
    (Transformer mean over three seeds). The iteration sweep
    ($\beta_\text{model}{=}0.3$, solid) and weight sweep
    (iter$_5$, dashed) intersect at the shared main configuration.
    Dotted lines mark the pure-DP reference, and the shaded region contains configurations that improve both metrics over this baseline. Subsequent refinement
  increases learnability while reducing segment-level separation. Pseudo-labeling (\texttt{×}) lies outside the structured DDSL trajectories.}
% Figure description: Scatter plot of segment-level Fisher Ratio against Transformer macro F1 on SSE50, showing the iteration sweep and the feedback-weight sweep as two connected paths that meet at the main configuration, dotted lines marking the pure-DP reference, and a separate pseudo-labeling point with the highest F1 and lowest Fisher Ratio.
  \label{fig:pareto}
\end{figure}

\paragraph{Two routes through the tradeoff.}
Figure~\ref{fig:pareto} places the iteration and $\beta_\text{model}$
sweeps in the same learnability--discriminability space on SSE50. The trajectories
intersect at the main configuration because iter$_5$ with
$\beta_\text{model}{=}0.3$ is shared by both experiments. Both routes
initially enter the region in which macro F1 and $\text{FR}_\text{seg}$
exceed the pure-DP baseline: at iter$_1$ along the iteration sweep and at
$\beta_\text{model}{=}0.1$ along the weight sweep. Beyond these initial
corrections, additional learnability is obtained at progressively lower
segment-level separation. The figure thus provides an empirical map for
selecting the number of iterations and feedback weight according to the
relative importance assigned to label recoverability and price-feature
separation. The two controls are not mechanistically interchangeable:
iteration approaches a fixed point under a fixed energy balance, whereas
$\beta_\text{model}$ changes that balance.
% ==============================================================
\subsection{External Validation: A Non-Anticipative Event Study}
\label{sec:event_study}
% ==============================================================
Macro F1 and $\text{FR}_\text{seg}$ are internal properties of the constructed
labels. We therefore test whether predictions learned from them retain
information about subsequently realized prices, checking that increasing
learnability does not make the labels economically empty.

\paragraph{Design.}
An event occurs when the predicted future state changes at day $t$,
$\hat s_{f,t}\neq\hat s_{f,t-1}$, where $\hat s_t$ is the joint state
emitted by the online Viterbi decoder, so each event is typed
$\hat s_{c,t}\rightarrow\hat s_{f,t}$. Observations are (symbol, day) pairs.
For each event we record the realized cumulative log return over the
following 20 trading days from adjusted close prices, and compare each
transition with a persistence control sharing the same current state. For
example, BEAR$\rightarrow$BULL is compared with BEAR$\rightarrow$BEAR,
isolating the contribution of the predicted future state at fixed current
state. We compute the difference in group means for each seed, in percentage
points, and report the average across seeds:
\begin{equation}
\Delta = 100\big(\overline{r_{20}}|_{\text{event}}
- \overline{r_{20}}|_{\text{ctrl}}\big),
\quad r_{20}=\log(C_{t+20}/C_{t}).
\label{eq:event_delta}
\end{equation}

Table~\ref{tab:event_study} reports the mean and standard deviation of
$\Delta$ across seeds. These are raw rather than abnormal returns: no expected-return model is estimated, and the same-current-state control subtracts only the
corresponding control-group mean. The groups are not additionally matched
on volatility, size, or prior return. Held-out labels are
never used as outcomes; only model predictions and subsequently realized
prices enter the analysis. Because event windows overlap and same-day events
are cross-sectionally correlated, we report descriptive effect sizes rather
than i.i.d.\ significance tests.

\paragraph{Online decoder.}
Predicted states are decoded from the test-period outputs of the
iter$_k$ Transformer. $E_\text{price}$ cannot be used online because it
reads prices after day $t$ (the future window and the right half of the
current window). The online emission is therefore the classifier's
log-probability alone,
$\hat e_t(s)=\log p_{\theta_k}(s\mid x_{t-T_\text{seq}+1:t})$, combined with
the labeling-stage grammar at unit weight:
\begin{equation}
\begin{aligned}
D_t(s)&=\hat e_t(s)+\max_{s'}\bigl[D_{t-1}(s')+E_\text{grammar}(s',s)\bigr],\\
\hat s_t&=\arg\max\nolimits_s D_t(s),
\end{aligned}
\label{eq:online}
\end{equation}
initialized with $D_{t_0}=\hat e_{t_0}$ on each stock's first test-period day.
Each $\hat s_t$ is emitted and frozen without backtracking, so it depends
only on inputs up to day $t$. The online rule thus differs from the
retrospective labeler in three ways: it has no price emission, it does not
backtrack, and the model term enters with unit weight rather than
$\beta_\text{model}$. No new hyperparameters are introduced.

\begin{table}[t]
\caption{Event-minus-control differences (\%) under non-anticipative online Viterbi
decoding, mean$\pm$std over three seeds. The two blocks isolate the same
predicted BULL call under different current states. Cells contain $3\times10^2$ to $8\times10^3$ events.}
\label{tab:event_study}
\centering
\small
\setlength{\tabcolsep}{4pt}
\begin{tabular}{llccc}
\toprule
\textbf{Transition} & \textbf{Dataset} & \textbf{iter$_0$} & \textbf{iter$_1$} & \textbf{iter$_5$} \\
\midrule
\multirow{4}{*}{BEAR$\to$BULL}
  & SSE50     & $+1.13\pm0.69$ & $+1.63\pm0.42$ & $+2.11\pm0.37$ \\
  & CSI300    & $+1.17\pm0.40$ & $+1.08\pm0.22$ & $+1.50\pm0.38$ \\
  & CSI500    & $+1.47\pm0.73$ & $+0.68\pm0.23$ & $+2.00\pm0.79$ \\
  & NASDAQ100 & $-1.51\pm0.50$ & $-1.65\pm1.15$ & $-2.32\pm0.75$ \\
\midrule
\multirow{4}{*}{NEUT$\to$BULL}
  & SSE50     & $+0.70\pm0.13$ & $+0.20\pm0.13$ & $-0.25\pm0.08$ \\
  & CSI300    & $+0.22\pm0.33$ & $+0.42\pm0.34$ & $+0.49\pm0.15$ \\
  & CSI500    & $+0.81\pm0.15$ & $+0.80\pm0.11$ & $+0.65\pm0.17$ \\
  & NASDAQ100 & $+0.05\pm0.12$ & $+0.09\pm0.32$ & $+0.03\pm0.12$ \\
\bottomrule
\end{tabular}
\end{table}
\paragraph{The structured state space remains informative.}
The grammar improves temporal coherence at prediction time: median
predicted-BULL duration is two sessions under daily argmax in nearly every
dataset--iteration cell, but five--six sessions at iter$_0$ and about three at
iter$_5$ under online Viterbi. The joint current--future state space also
enables conditional events that a future-only label cannot represent. In
Table~\ref{tab:event_study}, the divergence is clearest on SSE50: the same
predicted BULL call yields $+2.11\%$ at iter$_5$ under BEAR$\to$BULL but
$-0.25\%$ under NEUT$\to$BULL, whereas at iter$_0$ the two rows are much
closer ($+1.13\%$ and $+0.70\%$); each is measured against its own persistence
control. The current-state condition is therefore necessary to expose the
reversal signal, though the magnitude of this divergence varies across
markets.

\paragraph{Refinement preserves limited external value.}
In the top block, the BEAR$\rightarrow$BULL effect is positive for every
seed and iteration on all three A-share datasets and is larger at
iter$_5$ than at iter$_0$. Refinement therefore strengthens a conditional
bear-reversal signal despite the decline in $\text{FR}_\text{seg}$.
Transitions with future state BEAR (not shown) do not change consistently
across datasets and are not interpreted as systematic effects. On
NASDAQ100, the BEAR$\rightarrow$BULL effect is consistently negative and
becomes more negative. We have not tested an explanation for this sign reversal; differences in short-horizon reversal behavior between the two markets are one candidate. Thus, increasing learnability does not eliminate
the association between DDSL predictions and subsequently realized returns,
but the external evidence is event-specific and market-dependent rather
than a general validation of directional accuracy or trading profitability.

% ==============================================================
\section{Conclusion}
% ==============================================================

We presented DDSL, an iterative self-labeling framework that couples a
price- and grammar-guided Viterbi labeler with a Transformer through model
feedback, producing nine joint current--future market states. Across four
datasets and three seeds, refinement improves held-out macro F1 by
$78\%$--$96\%$ over single-pass DP labeling. An independent GRU reproduces
the trend, while label entropy and class coverage remain stable, supporting
cross-architecture learnability rather than only Transformer
self-\hspace{0pt}confirmation, and showing that the gain is not caused by label
collapse.

We further characterized how iteration and $\beta_\text{model}$ control this
gain. At fixed $\beta_\text{model}$, most improvement occurs in the first two
rounds before successive labelings converge and marginal gains diminish.
Increasing $\beta_\text{model}$ raises learnability further, but the widening
Transformer--GRU gap shows that stronger feedback produces increasingly architecture-specific gains. After an initial joint improvement, continued
refinement also reduces segment-level price-feature separation. The resulting
learnability--discriminability map therefore supports application-dependent
selection of iteration count and feedback strength rather than maximizing
Transformer F1 alone.

External validation provides limited but complementary evidence. Under
non-anticipative online Viterbi decoding, refinement strengthens the
conditional BEAR$\rightarrow$BULL reversal effect on all three A-share
datasets, whereas NASDAQ100 exhibits an association of the opposite sign.
The external value of DDSL is therefore event-specific and market-dependent,
rather than evidence of uniform directional accuracy or trading
profitability. The event study also shows that the same predicted BULL call
is associated with different subsequent returns depending on the predicted
current state, most clearly on SSE50, so the joint current--future
representation is needed to expose this conditional signal.

Several limitations remain. Test-split labels at iter$_k$ depend on the
iter$_{k-1}$ Transformer's outputs on the same test observations, so
held-out F1 measures the recoverability of a label system partly shaped by
that model. The GRU probe reduces but does not remove this dependence. The $\beta_\text{model}$ sweep covers only SSE50 and
NASDAQ100, and the datasets use a fixed December~2025 constituent snapshot
rather than point-in-time index membership. Further validation should
include a future period withheld from the refinement loop, isolated
cross-timestep grammar ablations, comparisons with simpler state spaces,
and sensitivity analysis of the initial labeler. Future work will develop
adaptive stopping and feedback schedules that balance learnability, label
structure, and non-anticipative financial evidence, and evaluate predicted
states as conditioning variables in downstream allocation and timing models.

\section*{Acknowledgments}
This work was supported in part through the NYU Shanghai IT High
Performance Computing resources, services, and staff expertise.

\bibliographystyle{ACM-Reference-Format}
\bibliography{ddsl_references}

@article{bellman1961approximation,
  title={On the approximation of curves by line segments using dynamic programming},
  author={Bellman, Richard},
  journal={Communications of the ACM},
  volume={4},
  number={6},
  pages={284},
  year={1961}
}

@article{hamilton1990analysis,
  title={Analysis of time series subject to changes in regime},
  author={Hamilton, James D},
  journal={Journal of Econometrics},
  volume={45},
  number={1-2},
  pages={39--70},
  year={1990}
}

@inproceedings{wu2021optimal,
  title={Optimal segmented linear regression for financial time series segmentation},
  author={Wu, Chi-Jen and Zeng, Wei-Sheng and Ho, Jan-Ming},
  booktitle={2021 International Conference on Data Mining Workshops (ICDMW)},
  year={2021},
  organization={IEEE}
}

@inproceedings{lee2013pseudo,
  title={Pseudo-label: The simple and efficient semi-supervised learning method for deep neural networks},
  author={Lee, Dong-Hyun},
  booktitle={Workshop on Challenges in Representation Learning, ICML},
  year={2013}
}

@article{fisher1936use,
  title={The use of multiple measurements in taxonomic problems},
  author={Fisher, Ronald A},
  journal={Annals of Eugenics},
  volume={7},
  number={2},
  pages={179--188},
  year={1936}
}

@inproceedings{vaswani2017attention,
  title={Attention is all you need},
  author={Vaswani, Ashish and Shazeer, Noam and Parmar, Niki and Uszkoreit, Jakob and Jones, Llion and Gomez, Aidan N and Kaiser, {\L}ukasz and Polosukhin, Illia},
  booktitle={Advances in Neural Information Processing Systems},
  volume={30},
  year={2017}
}

@article{zegadlo2022identifying,
  title={Identifying bull and bear market regimes with a robust rule-based method},
  author={Zegad{\l}o, Piotr},
  journal={Research in International Business and Finance},
  volume={60},
  pages={101603},
  year={2022},
  publisher={Elsevier}
}

@book{zakamulin2017market,
  title={Market Timing with Moving Averages: The Anatomy and Performance of Trading Rules},
  author={Zakamulin, Valeriy},
  series={New Developments in Quantitative Trading and Investment},
  publisher={Palgrave Macmillan},
  address={Cham},
  year={2017},
  doi={10.1007/978-3-319-60970-6}
}

@article{sohn2020fixmatch,
  title={{FixMatch}: Simplifying semi-supervised learning with consistency and confidence},
  author={Sohn, Kihyuk and Berthelot, David and Carlini, Nicholas and Zhang, Zizhao and Zhang, Han and Raffel, Colin and Cubuk, Ekin Dogus and Kurakin, Alexey and Li, Chun-Liang},
  journal={Advances in Neural Information Processing Systems},
  volume={33},
  pages={596--608},
  year={2020}
}

@inproceedings{tanaka2018joint,
  title={Joint optimization framework for learning with noisy labels},
  author={Tanaka, Daiki and Ikami, Daiki and Yamasaki, Toshihiko and Aizawa, Kiyoharu},
  booktitle={Proceedings of the IEEE conference on computer vision and pattern recognition},
  pages={5552--5560},
  year={2018}
}

@article{reed2014training,
  title={Training deep neural networks on noisy labels with bootstrapping},
  author={Reed, Scott and Lee, Honglak and Anguelov, Dragomir and Szegedy, Christian and Erhan, Dumitru and Rabinovich, Andrew},
  journal={arXiv preprint arXiv:1412.6596},
  year={2014}
}

@inproceedings{yarowsky1995unsupervised,
  title={Unsupervised word sense disambiguation rivaling supervised methods},
  author={Yarowsky, David},
  booktitle={33rd annual meeting of the association for computational linguistics},
  pages={189--196},
  year={1995}
}

@inproceedings{lin2017focal,
  title={Focal loss for dense object detection},
  author={Lin, Tsung-Yi and Goyal, Priya and Girshick, Ross and He, Kaiming and Doll{\'a}r, Piotr},
  booktitle={Proceedings of the IEEE international conference on computer vision},
  pages={2980--2988},
  year={2017}
}

@inproceedings{ma2022denoised,
  author    = {Ma, Yanqing and Ventre, Carmine and Polukarov, Maria},
  title     = {Denoised Labels for Financial Time Series Data via Self-Supervised Learning},
  booktitle = {Proceedings of the Third ACM International Conference on AI in Finance},
  series    = {ICAIF '22},
  year      = {2022},
  pages     = {471--479},
  publisher = {ACM}
}

@article{procacci2019forecasting,
  author  = {Procacci, Pier Francesco and Aste, Tomaso},
  title   = {Forecasting Market States},
  journal = {Quantitative Finance},
  volume  = {19},
  number  = {9},
  pages   = {1491--1498},
  year    = {2019},
  doi     = {10.1080/14697688.2019.1622313}
}

@inproceedings{lafferty2001conditional,
  title={Conditional random fields: Probabilistic models for segmenting and labeling sequence data},
  author={Lafferty, John and McCallum, Andrew and Pereira, Fernando C. N.},
  booktitle={Proceedings of the 18th International Conference on Machine Learning},
  pages={282--289},
  year={2001}
}

@book{lopezdeprado2018advances,
  title={Advances in Financial Machine Learning},
  author={L{\'o}pez de Prado, Marcos},
  year={2018},
  publisher={John Wiley \& Sons}
}

@article{kovacevic2023trend,
  title={Optimal Trend Labeling in Financial Time Series},
  author={Kova{\v c}evi{\'c}, Tomislav and Mer{\'c}ep, Andro and Begu{\v s}i{\'c}, Stjepan and Kostanj{\v c}ar, Zvonko},
  journal={IEEE Access},
  volume={11},
  pages={83822--83832},
  year={2023},
  publisher={IEEE},
  doi={10.1109/ACCESS.2023.3303283}
}

@article{hanna2018top,
  title={A top-down approach to identifying bull and bear market states},
  author={Hanna, Alan J},
  journal={International Review of Financial Analysis},
  volume={55},
  pages={93--110},
  year={2018},
  publisher={Elsevier}
}

@article{viterbi1967error,
  title={Error bounds for convolutional codes and an asymptotically optimum decoding algorithm},
  author={Viterbi, Andrew J},
  journal={IEEE Transactions on Information Theory},
  volume={13},
  number={2},
  pages={260--269},
  year={1967},
  publisher={IEEE}
}
\end{document}